\documentclass[conference]{IEEEtran}
\usepackage[version=4]{mhchem}
\usepackage{chemformula}
\usepackage{cite}
\usepackage{amsmath,amssymb,amsfonts}
\usepackage{algorithmic}
\usepackage{graphicx}
\usepackage{textcomp}
\usepackage{xcolor}

\usepackage{svg}
\usepackage{amsmath}
\usepackage{booktabs}
\usepackage{multirow}
\usepackage[flushleft]{threeparttable}
\usepackage{float}
\usepackage{tabularx}
\usepackage{footnote}    
\usepackage[caption=false,font=footnotesize]{subfig}
\definecolor{mygreen}{RGB}{137, 172, 107}

\ifCLASSINFOpdf
\else
\fi
\begin{document}

%
\title{Deep Multi-attribute Graph Representation Learning on Protein Structures}

\author{\IEEEauthorblockN{Tian Xia}
\IEEEauthorblockA{Department of Computer Science\\ and Software Engineering\\
 Auburn University\\
 Auburn, AL, 36849\\
Email: tianxia@auburn.edu}
\and
\IEEEauthorblockN{Wei-Shinn Ku}
\IEEEauthorblockA{Department of Computer Science\\ and Software Engineering\\
 Auburn University\\
 Auburn, AL, 36849\\
Email: weishinn@auburn.edu}}


%


\maketitle

\begin{abstract}
 Graphs as a type of data structure have recently attracted significant attention. Representation learning of geometric graphs has achieved great success in many fields including molecular, social, and financial networks. It is natural to present proteins as graphs in which nodes represent the residues and edges represent the pairwise interactions between residues. However, 3D protein structures have rarely been studied as graphs directly. The challenges include: 1) Proteins are complex macromolecules composed of thousands of atoms making them much harder to model than micro-molecules. 2) Capturing the long range pairwise relations for protein structure modeling remains under-explored. 3) Few studies have focused on learning the different attributes of proteins together. To address the above challenges, we propose a new graph neural network architecture to represent the proteins as 3D graphs and predict both distance geometric graph representation and dihedral geometric graph representation together. This gives a significant advantage because this network opens a new path from the sequence to structure. We conducted extensive experiments on four different datasets and demonstrated the effectiveness of the proposed method.

\end{abstract}



%
\IEEEpeerreviewmaketitle

\section{Introduction}
Prediction of a protein's structure from its amino acid sequence remains an open problem in the field of life science. The main practical problem confronting us is the challenge that comes from directly predicting protein structure from primary sequence. 
A common strategy used to study protein structure is to transform the direct prediction of protein structure into several problems, including contact map prediction, secondary structure prediction, torsion angles prediction and others. Especially with the recent growth of convolutional neural networks (CNNs), several convolution neural networks were proposed to tackle the problem in this field, such as contact map prediction\cite{Wang2017,Senior2020ImprovedLearning}, torsion angle prediction \cite{gao2018}, and protein structure-property prediction \cite{property2016}.
\begin{figure}[htbp]
\centerline{\includegraphics[width=\linewidth,height=\textheight,keepaspectratio]{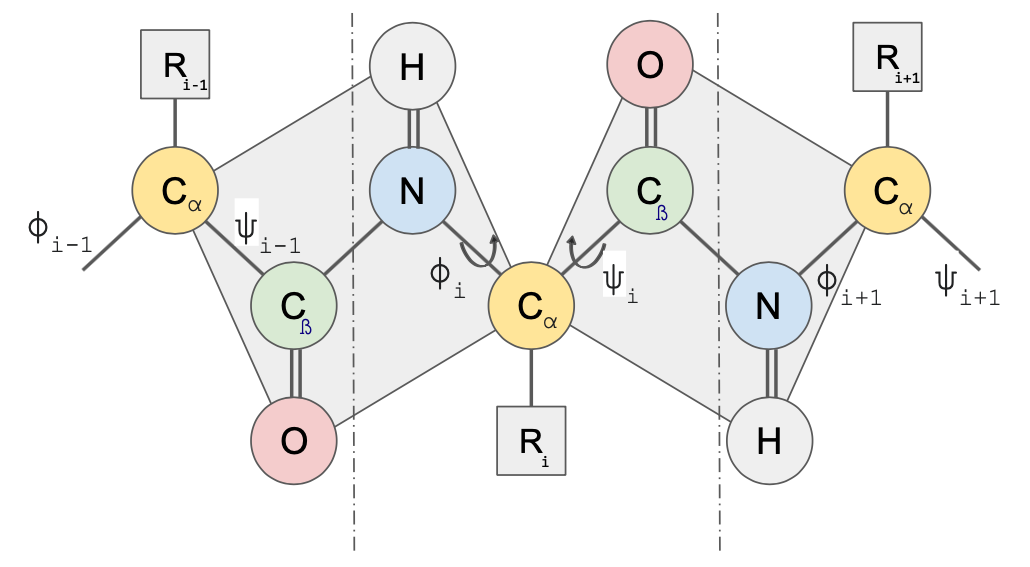}}
\caption{ Protein backbone representation. }
\label{protein_backbone}
\end{figure}

\begin{figure}[!h]
\centering
\subfloat[Details of image convolution.]{\includegraphics[width=2.3in]{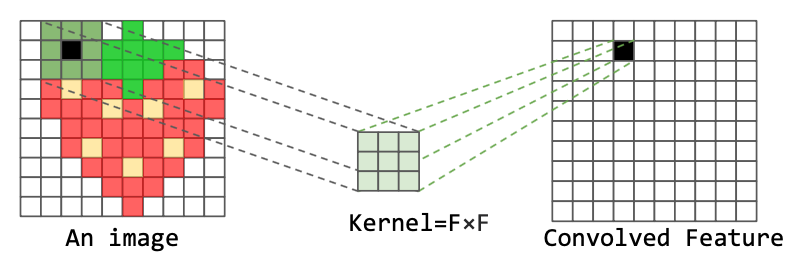}}
\hfil
\subfloat[Pairwise matrix]{\includegraphics[width=0.8in]{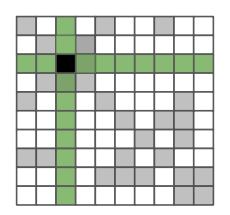}}
\caption{Illustration of image and pairwise distance matrix}
\label{image_conv}
\end{figure}

Following this road, a number of works have been done using deep convolution neural networks to predict the contact map first, then recover the protein structure from contact maps. Several challenges remain to be tackled:
\textbf{1) The foremost challenge is the fact that the protein contact/distance matrix cannot provide fully structural information needed for protein backbone structure modeling.} 
The contact map is a representation of contacts between residue pairs. The pairs of residues are considered in contact if the spatial distance of alpha carbons in the residues are below the predefined threshold. Contact maps are sparse matrices \cite{gdfuzz3d} that lack sufficient information to be treated by geometric representation because the only information that contact maps provide is the upper or lower bound of the contact threshold.
Multiple conformations can be generated since alpha carbon can move freely in the space within the contact threshold. Even though several recent studies \cite{badri2020, Xu16856} use the distance matrix instead of contact map which contains finer-grained information related to the alpha carbon pairwise distances than the contact map, they are still insufficient to model protein backbone geometry structure because of the absence of structural information of other backbone atoms in the protein. The free rotation of the chemical bonds around alpha carbon \cite{RAMACHANDRAN196395} cause the backbone atoms to rotate accordingly. Hence further constraint information is needed to model all atoms in the protein backbone structure.
As illustrated in Fig. \ref{protein_backbone}, the protein backbone structure composed of consecutive chains of coplanar units of \ce{C_\alpha -CO - NH - C_\alpha} with two primary degrees of freedom: dihedral angles named Phi ($\phi$) and Psi ($\psi$).
The rotation information on both sides of the \ce{C_\alpha} in addition to distance matrix could provide further geometric information of all the atoms in the protein backbone structure.
Although both distance information and dihedral angle information are helpful to study the protein backbone structure, few studies have focus on modeling multiple attributes for protein backbone structural representations.  
\textbf{2) Secondly, it remains challenging to capture the non-local relations in protein structures.}
Existing works\cite{Wang2017, Senior2020ImprovedLearning} typically use image CNN-based methods which focus on local neighbor amino acids in the protein sequence but cannot consider those far away in the sequence. As shown in Fig. \ref{image_conv}(a), the 2D image convolution operation only focuses on local information by multiplying the convolutional kernel over each pixel and its local neighbors. Although this operation is enough for image processing, one limitation of these methods is that our pairwise feature matrix is not image representation. 
Fig. \ref{image_conv}(b) shows an example of the pairwise relation matrix, which represents the amino acid pairs in the feature representation space. After folding, amino acids sequentially far apart may be folded into close spatial regions.
Hence, we need not only consider the local relations but also other pairwise relations denoted in the column and row in Fig. \ref{image_conv}(b). The green cross in Fig. \ref{image_conv}(b) shows the edges connected to the same nodes for the black square edge. Although all the pairwise features in the green cross could potentially influence the pairwise relation in the black square, only the local relations were considered in the image convolution process.
Thus it remains challenging to model long-range relations in the protein sequences. 
3) \textbf{The variable protein sequence lengths and the large size of the protein structures make the problem difficult to handle}. This problem is usually overcome by transforming the 2D feature maps via cropping \cite{Senior2020ImprovedLearning, billings2019prospr} or padding\cite{Eguchi2020.08.07.242347}. Although such operations generate fixed-sized inputs for the neural network, they could potentially cause the loss of relational details or undesired distortion of the contact/distance matrix. While there are few studies which model the chemical molecules as a graph, considering protein as a whole graph to study its structural properties remains an open problem in the area. The long length of protein sequences and the large sizes of the proteins have limited the approaches to model protein as one graph directly.

To rectify the above problems, we investigate the native structures of the protein and their common representations. Although the natural way to represent a protein structure is to model it as a 3D graph, the protein 3D graph structure has rarely been studied directly. Recent theoretical developments in graph neural networks inspired us to look at protein structure representation differently. 
Fig. \ref{protein_graph} shows the structure of protein 2XSE. The backbone of the protein holds a protein structure together with residues of each amino acid. The \ce{C_\alpha} is the central atom in the backbone structure, which has two backbone angles $(\phi, \psi)$. Dihedral angles $(\phi, \psi)$ for alanine in protein 2XSE are marked in Fig. \ref{protein_graph}. 
Although the contact map was widely considered to be a good way to represent protein structure, we focus on overcoming the protein structure problem by modeling the protein structure as a 3D geometric graph. 
We designed a geometric graph convolutional network architecture based on this specific problem. Our goal is to generate distance geometric graph representation and dihedral geometric graph representations together for protein structure modeling. Compared to positional 3D graph representations, i.e., Cartesian coordinates of nodes, our proposed 3D graph representation method gives a significant advantage because of its invariance to rotation and translation of the graph. To the best of our knowledge, this is the first work that can address all the above challenges. We summarize the main contributions as follows:

\begin{figure}[htbp]
\centerline{\includegraphics[width=\linewidth,height=\textheight,keepaspectratio]{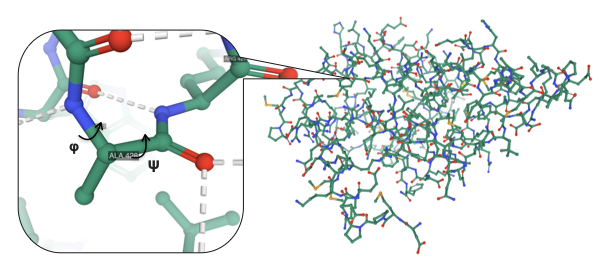}}
\caption{Graph representation of the protein structure. }
\label{protein_graph}
\end{figure}

\begin{itemize}
    \item We formally formulate the problem of protein backbone structure modeling as geometric 3D graph representations. We model the input graph into multi-attribute graphs in which the node represents the residues and the edge represents pairwise information between residues. 
    \item A novel architecture is proposed for protein backbone 3D structure graph generation. Our proposed model could generate a protein graph with both geometric distance graph representations and geometric dihedral graph representations together. 
    \item We propose the use of novel geometric graph convolution blocks for distance geometric graph representation generations. As the sizes of the protein vary, our proposed approach can handle the sizes of protein graphs dynamically. 
    \item Comprehensive experiments were conducted to validate the effectiveness of our proposed model in the generated 3D geometric protein graphs.
\end{itemize}
The rest of the paper is organized as follows. We review the related work in Section II, then formulate the problem in Section III. Section IV details the preliminaries and graph geometric representations. Section V presents the proposed network. Sections VI and VII report the experiments and results. Lastly, Section VIII concludes the paper and discusses the directions for the future work. 

\section{Related Work}
In this section, we will present and discuss three lines of research that are relevant to our work. 

\subsection{Protein structure prediction}
Experiments for protein structure determination are time consuming and expensive; thus, modelling the 3D structure of a protein remains one of the most important problems in bioinformatics\cite{Li2020ProteinDenseNet}. Significant work has been done toward the construction of the protein 3D structure during recent decades. Critical Assessment of Protein Structure Prediction (CASP) \cite{CASP} established benchmarks and assessed methods for protein structure prediction. Protein contact map and conformation prediction have proven to help the reconstruction of protein 3D structure\cite{Pietal2015, Eguchi2020.08.07.242347}. \cite{Yang1496, gao2018} showed that inter-residue orientations in addition to residue distances can be used for protein structure prediction \cite{AlQuraishi2019, Eguchi2020.08.07.242347}.
\subsection{Deep Learning for Protein Structure Prediction}

Deep learning-based methods were already widely used to solve protein structure related problems, such as protein-protein interaction prediction, protein contact map prediction, protein secondary structure predictions, and protein dihedral angle predictions. Deep learning methods for inter-residue distance and contact prediction have considerably advanced protein structure prediction. \cite{Du2020Energy-basedConformations} proposed an energy-based model using transformer architecture for protein conformation. \cite{alquraishi2019end} proposed end-to-end recurrent geometric network to predict 3D protein structure. \cite{Anand2019} generated a protein 3D structure by using Generative Adversarial Networks.
Currently, the most successful methods for residue-residue distance prediction are CNN-based neural networks. 
\cite{Wang2017} first applied CNN to predict protein contacts.
\cite{Adhikari2020}, \cite{Fukuda2018}, \cite{Senior2020ImprovedLearning} and \cite{Wu2020a} used dilated CNN method for protein contact map prediction.

\subsection{Graph Neural Network}
Graph neural network (GNN) attracts attention in a wide range of areas\cite{zhou2018graph} including natural language processing, computer vision, traffic prediction, and so on. 
Graph convolutional network, which originated from spectral graph convolutional neural networks\cite{bruna2013spectral}, has shown practical utility in the field of chemistry. 
Inspired by GNN, several works treat molecules as graphs and achieved great progress. \cite{Mansimov2019MolecularNetwork} used graph neural networks to learn energy function for small molecules. \cite{Cho2018Three-DimensionallyInterpretation} proposed a three-dimensional graph convolution network to predict molecular properties. \cite{Gligorijevic2019Structure-BasedNetworks} uses graph convolutional networks to predict protein site-specific functions. \cite{klicpera2020directional} also utilizes GNN to tackle the molecule properties prediction problem.

\section{Problem Formulation}
The protein backbone holds the protein together and generates the tertiary structure of the protein. This section introduces the protein backbone geometry problem: this is a sequence to structure task where we take protein amino acid sequences as input to predict the geometry of 3D protein backbone structure. To tackle this complex problem, we first formulated the protein structure into a graph representation as follows:

We consider each input protein as a graph $G(\mathcal{V},\mathcal{E},E,F)$, where $\mathcal{V}$ is the set of $L$ nodes in the graph representing amino acid residues and $\mathcal{E}\subseteq \mathcal{V} \times \mathcal{V}$ is the set of $L-1$ edges in the protein sequence. $F\in \mathbb{R}^{L \times D}$ stands for the input node attribute matrix where $F_i\in \mathbb{R}^{1 \times D}$ refers to the node attribute of node $i$ and $D$ is the dimension of the node attribute vector. The input node attributes $F$  include position-specific scoring matrix, predicted secondary structure, solvent accessibility, etc.  $E\in \mathbb{R}^{L \times L \times K}$ is the edge attributes tensor, where $E_{i,j}\in \mathbb{R}^{1 \times K}$ refers to the edge attribute of edge $e_{i,j}$ and $K$ is the dimension of edge attributes tensor. Likewise, the input edge attributes tensor $E$ include co-evolution information, distance potential, and inter-residue coupling score.

Similarly, the target protein backbone geometry graph can be represented as $G(\mathcal{V'},\mathcal{E'},E',F')$, where $\mathcal{V'}$ is the set of $L$ nodes in the graph representing target amino acid residues and  $\mathcal{E'}\subseteq \mathcal{V'} \times \mathcal{V'}$ is the set of $M$ edges in the target graph. 
$E'\in \mathbb{R}^{K\times L \times L}$ denotes as the target edge attribute matrix where $E'_{k,i,j}$ denotes the $k$-th feature of edge $e_{i,j}$. $F'\in\mathbb{R}^{H\times L}$ denotes as the target node attribute matrix, where $F'_{k,i}\in\mathbb{R}^{1\times H}$ is the node attributes of node $i$ and $H$ is the dimension of the node attributes. For example, the $k$-th node attribute phi of node $i$ is the dihedral angle between the plane \ce{C^{i-1}_\beta, N^i, C^i_\alpha} and the plane \ce{N^i, C^i_\alpha, C^i_\beta} as shown in Fig. \ref{protein_angle}(a). In addition, psi is also a node feature which represents the dihedral angle between the plane \ce{N^i, C^i_\alpha, C^{i}_\beta} and the plane \ce{C^i_\alpha, C^i_\beta, N^{i+1}} as shown in Fig. \ref{protein_angle}(b). Without loss of generality, we assign the first node feature for torsion angle phi, namely $F'_{1}$ and the second node feature for torsion angle psi, namely $F'_{2}$.

Our goal is to develop a graph translation model that can encode both the node and edge features extracted from input protein graph $G(\mathcal{V},\mathcal{E},E,F)$ and generate a graph-based geometric representation for protein 3D structures $G(\mathcal{V},\mathcal{E'},E',F')$. 
As the input graph and output graph have the same protein sequence composed of the same set of nodes, we have $\mathcal{V} = \mathcal{V'}$. 
\begin{figure*}[htbp]
\centerline{\includegraphics[width=\linewidth,height=\textheight,keepaspectratio]{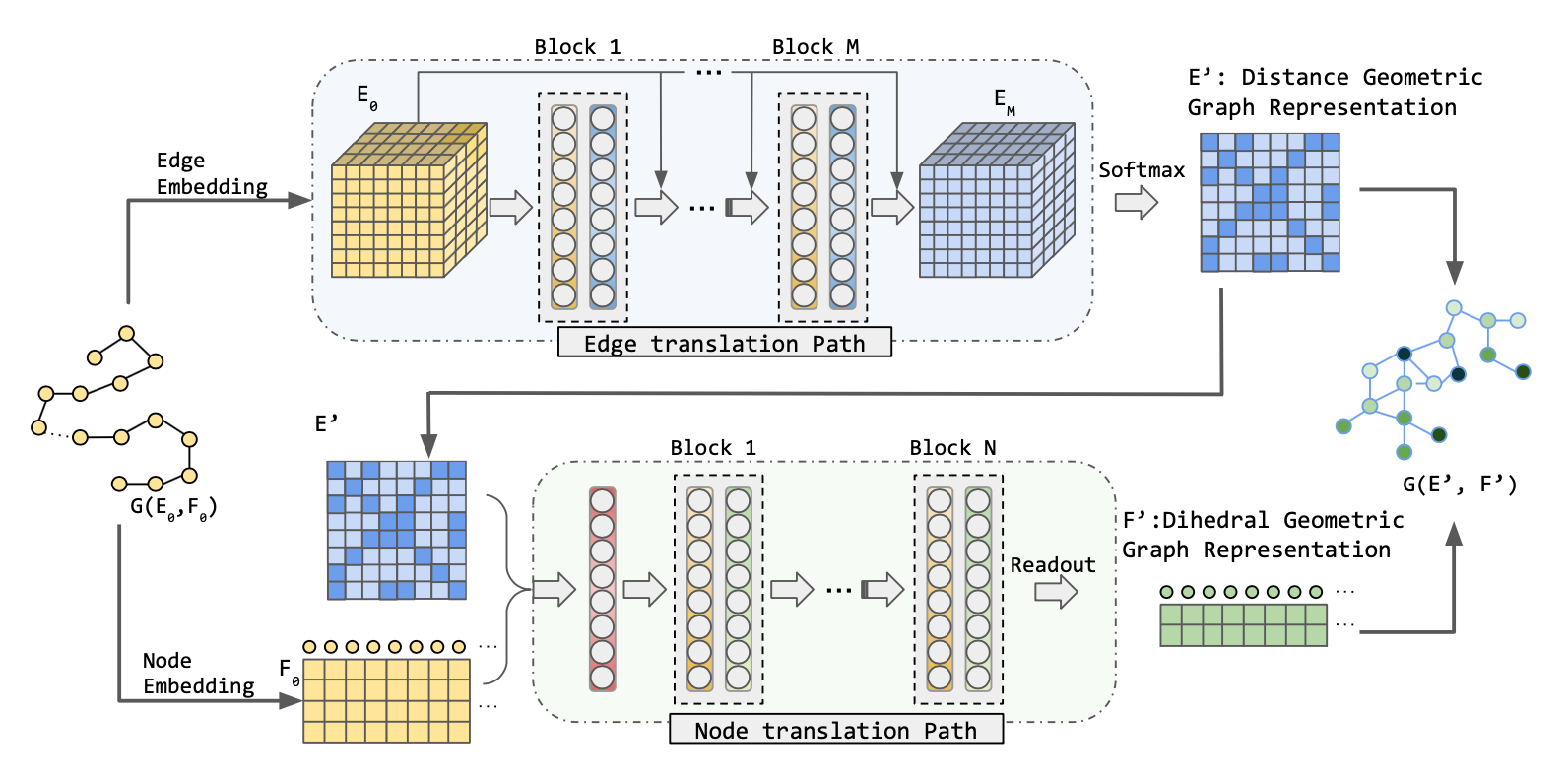}}
\caption{Overall architecture. The framework consists of two parts: the edge translation path is shown on top while node translation path is shown on the bottom. Given input protein sequences, we extract the features for both nodes and edges. Then the input can be denoted as $G(\mathcal{V},\mathcal{E},E,F)$ and fed into the framework as shown. }
\label{architecture}
\end{figure*} 
Hence the output is graph-based representation for protein backbone geometry $G(\mathcal{V},\mathcal{E'},E', F')$, where the pairwise distance information is represented by edge attributes $E'$ and the dihedral angle information is represented by node attributes $F'$. The main advantage of such geometric representation is the invariant property of angle-geometric graph representation and distance-geometric graph representation. Hence they are invariant under rotation of the coordinate system and graph translation. As most of the research in this field is aimed at getting only distance representation or torsion representation separately, here we solve the above mentioned problems simultaneously by 3D geometric graph generation. Since the input graph node and edge attributes $E,F$ and the target graph node and edge attributes $E', F'$ are different, the learning from the multi-attribute graph input to the graph geometric representation output can be defined as learning a mapping: $G(\mathcal{V},\mathcal{E},E,F) \rightarrow G(\mathcal{V},\mathcal{E'},E',F')$

\section{Preliminaries}

As illustrated in Fig. \ref{protein_backbone}, protein backbone structure consists of distance based geometric representation and dihedral angle representation. We proceed to detail the geometric representations of protein graphs. 

\subsection{Distance-based geometric graph representation}
We denoted edge attributes $E'$ as distance geometric representation for protein graph. 

\cite{levitt1975computer} introduced a chain of pseudoatoms placed at \ce{C_\alpha} positions to replace the protein main chain model. We adapted this simplification of the protein geometry and use distance matrix between \ce{C_\alpha} to represent the protein geometry. Thus, the distance matrix $E'$ of \ce{C_\alpha} determines the overall shape of the backbone structure. 

\begin{figure}[htbp]
\centerline{\includegraphics[width=\linewidth,height=\textheight,keepaspectratio]{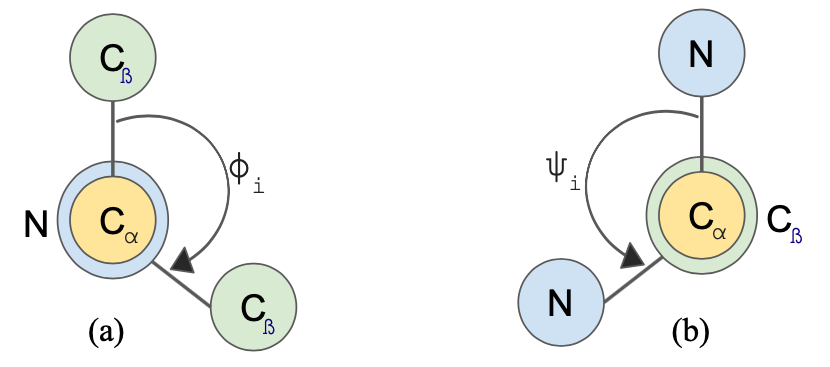}}
\caption{Dihedral angles Phi and Psi (a) The Phi $\phi$ torsion angle measures the rotation of the \ce{N-C_\alpha} bond (b) The Psi $\psi$ torsion angle measures the rotation of the \ce{C_\alpha-CO} bond.  }
\label{protein_angle}
\end{figure}
\subsection{Dihedral angle geometric graph representation}

The distance matrix $E'$ itself cannot fully characterize the overall backbone structure due to the rotation of chemical bonds around \ce{C_\alpha} : the \ce{N-C_\alpha} bond and the \ce{C_\alpha-CO} bond\cite{RAMACHANDRAN196395}. These two bonds attached to the  \ce{C_\alpha} can rotate freely to form the unique folding patterns within protein. As shown in Fig. \ref{protein_backbone}, the backbone dihedral angles Phi $(\phi)$ and Psi $(\psi)$ are in sequence order on either side of  \ce{C_\alpha} to represent the rotation of these bonds. Thus Phi $(\phi)$ and Psi $(\psi)$ can be denoted as node attributes  $F'$ to evaluate the rotation patterns' rise around  \ce{C_\alpha}. For $i^{th}$ amino acid, the dihedral angle $\phi_i$ is the dihedral angle between the plane \ce{C^{i-1}_\beta, N^i, C^i_\alpha} and the plane \ce{N^i, C^i_\alpha, C^i_\beta} as shown in Fig. \ref{protein_angle}(a). The dihedral angle $\psi_i$ is the dihedral angle between the plane \ce{N^i, C^i_\alpha, C^{i}_\beta} and the plane \ce{C^i_\alpha, C^i_\beta, N^{i+1}} as shown in Fig. \ref{protein_angle}(b). As each  \ce{C_\alpha} associates with two torsion agles $(\phi,\psi)$, we denote the dihedral angle geometric graph representation as node attributes matrix $F'_1, F'_2$. 
\section{Methodology}
In this section, we propose Protein Geometric Graph Neural Network (PG-GNN) to model the geometric properties in terms of distance and angle representations and learn the geometric representations jointly from two separate translation paths. We first describe the overall architecture of the PG-GNN with the translation paths. We then describe in detail how the geometric representations are learned with our proposed edge translation path and node translation path collaboratively. 
\begin{figure*}[!h]
\centering

\includegraphics[width=\linewidth]{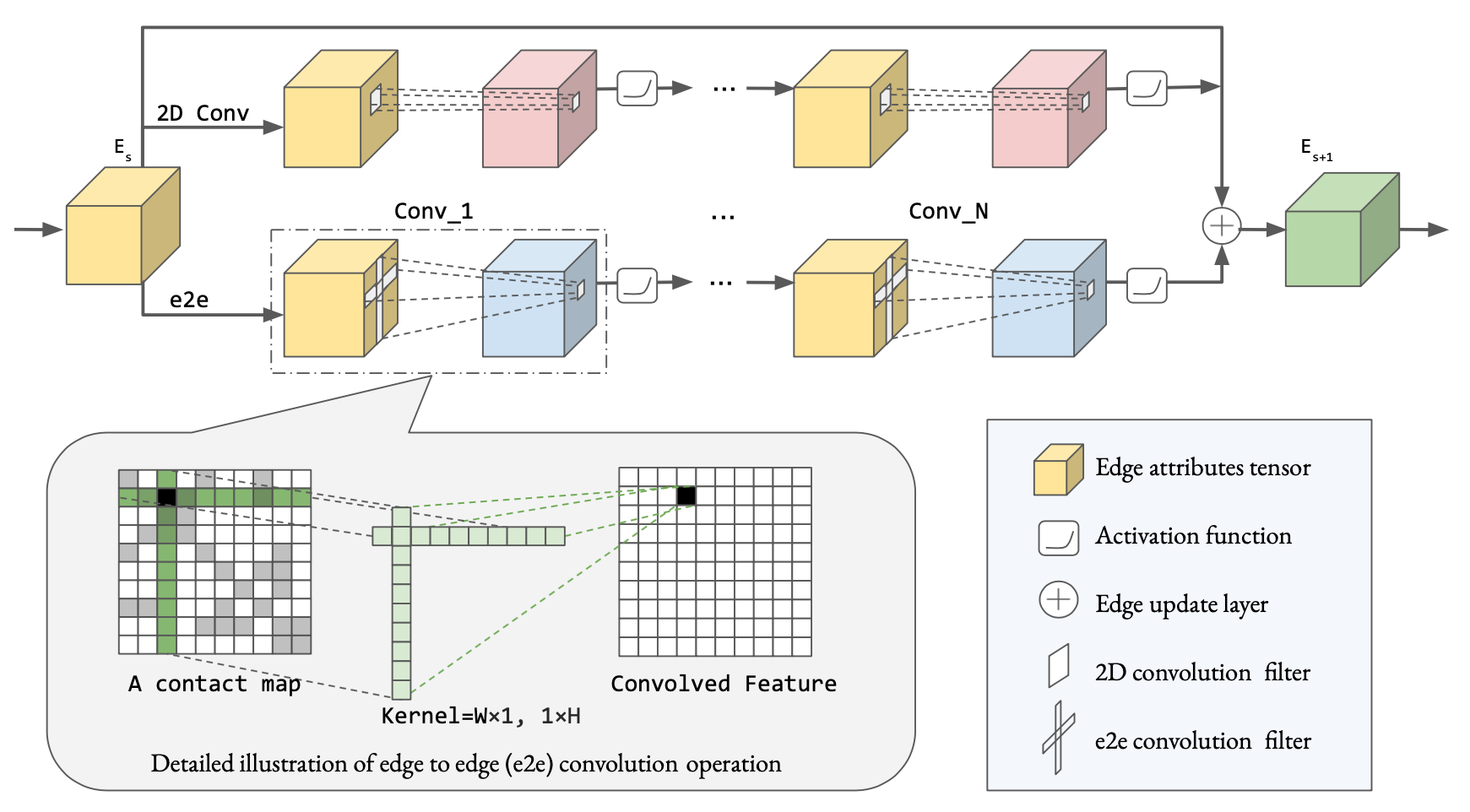}
\caption{Details of protein graph edge convolution path for one protein in a single block}
\label{e2e_conv}
\end{figure*}

\subsection{Model construction}

Taking protein sequences as input to directly construct geometric 3D representation of the protein structures remains an open problem in the area\cite{AlQuraishi2019}.
In light of the above discussion,  we need a framework that can dynamically handle different sized graph inputs and jointly generate output of both node and edge attributes together. With this aim in mind, in this paper we present a new framework composed of two translation paths to predict edge and node attributes separately.  The key components of our framework are edge translation path and node translation path. The illustration of our proposed framework is shown in Fig. \ref{architecture}. For the edge translation path, a deep residual convolutional network is proposed that takes both node and edge features as the input and output information on the pairwise distance of all residue pairs in the protein. The objective of the edge translation path is to learn the mapping: $G(\mathcal{V},\mathcal{E},E,F) \rightarrow G(E')$. For the node translation path, we utilize a fast graph message-passing neural netowrk which takes predicted pairwise distance potential and node features then outputs all node torsion angles $(\phi, \psi)$ in the protein graph. The objective of the node translation path is to learn the mapping: $G(\mathcal{V},\mathcal{E},E,F) \rightarrow G(F')$. The overall network is based on minimization with distance and orientation restraints derived from both edge translation path and node translation path outputs.

Although we can generate both node and edge attributes on separate paths based on the framework described above, the predicted attributes may not be consistent as they are generated from different paths.

For the edge translation path, we use cross entropy loss; only the probability corresponding to ground-truth participates in calculation. The calculation can be written as $ L({E,E'}) = -\frac{1}{L \times L}{\sum_{i=1}^{L}\sum_{j=1}^{L}{{y_{ij}}} \times \log \hat{y}_{ij} }$
where $L$ is the number of residues for each protein, $\hat{y}_{ij}$ is the predicted distance label between $i$-th residue and $j$-th residue in a protein.

For the node translation path, we use mean squared error as loss function.  The equation is as follows:
  \begin{multline}
  L(F,F') = L(F_1, F_1')+ L(F_2, F_2') \\
= \frac{1}{L}{\sum_{i=1}^{L}{\Big(\sin(\phi_i)-\sin(\hat{\phi_i})}\Big)^2} +  \frac{1 }{L}{\sum_{i=1}^{L}{\Big(\cos(\phi_i)-\cos(\hat{\phi_i})\Big)^2}} \\
+ \frac{1}{L}{\sum_{i=1}^{L}{\Big(\sin(\psi_i)-\sin(\hat{\psi_i})\Big)^2}} 
        +  \frac{1 }{L}{\sum_{i=1}^{L}{\Big(\cos(\psi_i)-\cos(\hat{\psi_i})\Big)^2}}
  \end{multline}
where $L$ is the number of residues for each protein.

For the overall training process, our PG-GNN is under the co-guidance of both edge translation path to learn the graph edge attributes and graph node translation path to learn the node attributes information. We set a parameter $\lambda$ to balance the degree of the two models. Thus the overall loss function for the network is:
\begin{equation}
 L =  L(E,E') + \lambda \times L(F,F')
\end{equation}

\subsection{Edge Translation Path (ETP)}
The aim for edge translation path is modeling all the interactions between edges and nodes to generate the geometric distance graph representation of the protein backbone structure. 
One challenging problem for 3D protein structure modeling is that residues sequentially far apart might be in spatially close contact in protein's 3D structure. Based on the protein sequence separation and the residue contact distance, there are different types of non-local contacts between residues: long-range contacts, medium-range contacts, and short-range contacts. Short range contacts are defined as residues with sequence separation of 6-11 residues, medium range contacts are residues with sequence separation of 12-23 residues, and long range contacts are residues with sequence separation more than 24 residues.

Interactions from all contact ranges can influence edge attributes generation.  
Moreover, one residue might have both local and non-local contacts with multiple residues at the same time. The influences between the interactions should also be captured for edge attributes generation. 
Furthermore, the variable protein sequence lengths and the large size of the protein structures make the problem difficult to handle with the existing models. 

Designing a translation path that can model different range interactions between residues and effects of all interactions connected to one residue is the key for our edge translation path. As the image convolution kernel only focuses on the source and its surrounding pixels, the long range patterns cannot be characterized by traditional image convolution kernels. Thus how to model both of the non-local and local pairwise relations in feature representation space becomes the crucial factor for protein structure modeling. Therefore, we propose a multi-branch convolution block for edge translation path to capture all the influences from different interaction types in proteins. The proposed edge translation block is shown in Fig. \ref{e2e_conv}. Each block is a multi-branch block with one identity mapping branch, one edge-to-edge convolution branch, and one two-dimensional (2D) convolution branch. The latter two branches used different hyper-parameters (filter type and sizes) separately. With 2D convolution operation focuses on the local relations and edge-to-edge convolution operation focuses on the long range contacts, our proposed edge translation block integrates both the local and non-local contacts features to generate each edge's attribute. Fig. \ref{e2e_conv} shows the operation branches in a single block, where the input graph attributes $E_s$ updated to output graph edge attributes $E_{s+1}$.

\textbf{Edge-to-edge convolution layers}
We implement an edge-to-edge filter which can consider the edges that connect to one residue. As shown in Fig. \ref{e2e_conv}, the protein contact map has its distinct features, unlike images. Thus, simply integrating the image convolution methods without modification will not work well for this problem. The edge-to-edge convolution is described as follows.

Formally, we denote $A \in \mathbb{R}^{L \times L}$ as an adjacency matrix where $L$ is the length of each protein, $W^{i,j}$ as the shared weights for the edge $e_{i,j}$, and  $\rho$ as the non-linear differentiable function that computes the activations. The graph convolution operation over the edges $e_{i,j}$ for kernel $W$ can be defined with $f_W^{i,j} = W^{i,j} \cdot  A_{i,j}^{l-1,m}$, likewise  the convolution operation  $f_H^{j,i}$ for kernel $H$ can be defined as $f_H^{j,i} = W^{i,j} \cdot  A_{j,i}^{l-1,m}$. Then the edge-to-edge convolution is as follows:
\begin{equation}
    A_{i,j}^{l,n} = \rho \left( \sum_{n=1}^{L}(f_W^{i,j} + f_H^{j,i}) \right)
\end{equation}
For the edge $e_{i,j}$, the shared weights $W^{i,j}$ across kernel $H$ and $W$ make the size of the latent convolutional representation independent of the size of the input.

\textbf{2D convolution layers}
We used $3\times3$ convolutional filters for the 2D convolution layers, followed by batch normalization and Exponential Linear Unit (ELU) activation function. Ten mini-blocks were used in total.
Dilation convolution filters can also be used in this convolution branch. The illustration of the detailed convolution operation is shown in Fig. \ref{image_conv}(a). 

\textbf{Edge updating layer}
Each edge is updated by integrating both of the convolution operations outputs and identity mapping of the inputs.
All layer outputs and residual connections are added together and fed into the next block. The details are shown in Fig. \ref{e2e_conv} as $\oplus$ operation. In this way, the edge will be generated by both related nodes in the input sequences and edges that connected to the same nodes. 

\subsection{Node Translation Path}

For the node translation path, the model aims to generate the node attributes matrix $F'$  based on learning the interactions between protein residues and the effects of the residue edges to the residues. As shown in Fig. \ref{protein_backbone}, the torsion angles $(\phi, \psi)$ are influenced by both of the connected edges and nodes. Hence both node attributes and the output adjacent matrix of the edge translation path are used as the input for our node translation path. We use message passing to capture all the nodes and edges influences for the torsion attribute generation. The overall architecture of each block in node translation path contains two layers: message passing layer and node update layers. The message passing layer learns all the influences from each pair of nodes and the node update layers aggregate all the influences and generate the new node attributes. 

\textbf{Message Passing on node layers:}
As shown in Fig. \ref{architecture}, the input of the proposed node translation path is sampled from both the node representation $F$ and edge representation $E$. 
We take the output distance matrix from the edge translation path at each iteration to use as edge representation $E_{vw}$ to feed into the node translation network\cite{li2015gated}. The inputs are fed into the message-passing layer as Equation \eqref{mp} to aggregate all incoming messages.

\begin{equation}
M_t(h_v^t, h_w^t, e_{vw}) = A_{e_{vw}}h_w^t
\label{mp}
\end{equation}

\textbf{Node updating layers:} After computing the message passing of all nodes, we update the hidden state by the Gated Recurrent Units (GRU) as Equation \eqref{gru}. 

\begin{equation}
    h_i^{t+1} = GRU(h_i^t,m_i^{t+1})
    \label{gru}
\end{equation}

\textbf{Readout layer:} 
We denote node attributes  $F_i =(F_{1,i}, F_{2,i})$ for $i$-th residue in the protein sequence. We further represent the dihedral angles as $v_{i} = (v_{a,i}, v_{b,i}, v_{c,i}, v_{d,i}) $, which denote the $\sin{\phi_i}, \cos{\phi_i}, \sin{\psi_i}, \cos{\psi_i}$ for training purposes. The readout layer is shown in Fig. \ref{architecture} as R in the last layer of the node translation path. We have the readout function as follows:
\begin{equation}
    \hat{v} = R(\{h_v^T|v \in G \})
\end{equation}

Then we derive the prediction of dihedral angles of  $\hat{F_i} =( arctan(\hat{v}_{a,i}/\hat{v}_{b,i}),arctan(\hat{v}_{c,i}/ \hat{v}_{d,i}))$  from the output of the model $\hat{v}_{i} =(\hat{v}_{a,i}, \hat{v}_{b,i}, \hat{v}_{c,i}, \hat{v}_{d,i}) $. 

\section{Experiment}
In this section, we present our experiments using the proposed methods on five real-world test datasets.

\subsection{Dataset}
The datasets that are used in the experiments are all real-world datasets. The proteins with a sequence length of more than 300 residues were excluded from the datasets. The contact matrix for the proteins in the test sets was calculated using the C$\alpha$ atoms distance. We define the contact between two residues where the relative distances between C$\alpha$ atoms of given residues are less than 8Å.

Training/validation dataset:
We divided the PDB25\cite{Wang2017} dataset as training and validation dataset. The PDB25 dataset contains proteins where the maximum sequence identity that any two proteins can share was 25\%. The proteins were filtered as described in \cite{Wang2017}. After the filtering process, 500 proteins were randomly sampled for the test set, and the remaining proteins were used for the training and validation of the models. The total number of proteins used in the training dataset is 4726 proteins and the validation dataset is 500 proteins. The validation dataset is used for hyperparameter tuning and to prevent model overfitting.

Testing dataset:
The trained models were tested on the following datasets: the PDB25 dataset described above, the protein domains used in the CASP11 and CASP13 competitions, 76 hard CAMEO benchmark proteins\cite{Wang2017}, and a test set that contains 400 membrane proteins (membrane dataset).

\subsection{Multi-attributes graph representation}
Based on our formulation, each protein is represented by a graph $G(\mathcal{V},\mathcal{E},E,F)$ composed of nodes $V$ presenting residues and edges $E$ represented by the protein sequence. 
Thus the two types of features used in this project are named node features and edge features accordingly. Node feature matrix $F$ are the properties of the single residue, including position-specific scoring matrix\cite{Beckstette2006}, predicted secondary structure, and solvent accessibility predictions. Edge feature matrix $E$ are the features that contain pairwise information, such as co-evolution information\cite{ccmpred}, and distance potential\cite{betancourt1999pair},\cite{Wang2017}. The extracted node features and edge features are used for the model described above. For edge translation path, the node features $F_i$ and $F_j$ are transformed by 1D convolution then concatenated into $E_{i,j}$ and  $E_{j,i}$ to be used as feature map for edge attributes generation. For node translation path, the node feature matrix $F$ and the pairwise distance matrix $E'$ from the edge translation path output are used as the input for node attributes generation.

\subsection{Parameters}
The parameters used in PG-GNN are presented in this section. All experiments are conducted on a 64-bit machine with Nvidia GPU (RTX 2080 Ti). For edge translation path, the number of edge translation blocks $S_E = 10$. For each edge translation block, 4 convolution layers per block $N$ was applied sequentially. In parallel, one edge-to-edge layer was applied to input of the edge translation block. For both convolution and edge-to-edge translations ELU activation function was used after each convolution layer. For node translation path, the number of blocks in message passing is $S_V = 6$ with ReLU activation function in each block. The network was trained using Adam \cite{adam} with an initial learning rate of 0.00013.

\begin{figure*}[h!]
\centerline{\includegraphics[width=\linewidth,height=\textheight,keepaspectratio]{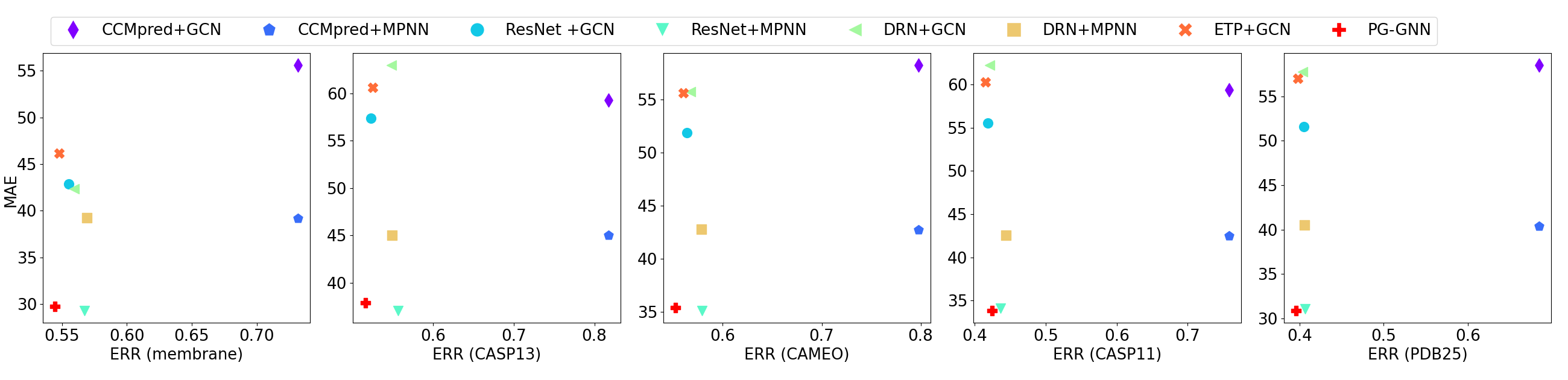}}
\caption{Overall performance comparison }
\label{result_comparison}
\end{figure*}

\begin{table*}[htbp]
  \centering
  \renewcommand{\arraystretch}{1.3}
  \caption{Performance comparison with baselines(PDB25 test dataset)}
    \begin{tabular}{lccccccccccccccc}
    \toprule
     \multirow{2}[2]{*}{\centering Metrics}  & \multicolumn{4}{c}{Edge LR-ACC} & \multicolumn{4}{c}{Edge MR-ACC}& \multicolumn{4}{c}{Edge SR-ACC}&\multicolumn{2}{c}{Node MAE}\\ 
   \cmidrule(lr){2-5}   \cmidrule(lr){6-9}  \cmidrule(lr){10-13}  \cmidrule{14-15} &  L/10  & L/5 &  L/2  & L & L/10&  L/5   &  L/2  &  L   & L/10 &L/5 & L/2  &   L&  Phi& Psi \\
        \midrule
CCMpred+GCN   & 0.537    &    0.473  &      0.354  &    0.249 & 0.466   &     0.360   &     0.224  &    0.149  &  0.379    &    0.289   &     0.181  &    0.124&  34.541     &   82.427    \\
CCMpred+MPNN  & 0.537 &       0.473  &      0.354  &    0.249 &  0.466   &     0.360  &      0.224  &    0.149 &   0.379    &    0.289    &    0.181  &    0.124&  24.957  &      55.833   \\
ResNet +GCN   &0.793    &    0.753  &      0.652    &  0.513 &0.789    &    0.698     &   0.492  &    0.313  & 0.792    &    0.670  &      0.425   &   0.249& 30.494  &      72.672    \\
ResNet+MPNN   &   0.794  &      0.753  &      0.646  &    0.507 & 0.786   &     0.693   &     0.488 &     0.312  &  0.794   &     0.671    &    0.424   &   0.251& 24.372    &    37.743       \\
DRN+GCN &0.795   &     0.757  &      0.654  &    0.511 &  0.789    &    0.696 &       0.494  &    0.316 &  0.788   &     0.674 &      0.428   &   0.252& 32.900  &      82.573   \\
DRN+MPNN   &  0.800   &     0.754     &   0.647  &    0.506 & 0.792  &      0.690   &     0.487  &    0.314  & 0.792 &       0.672  &      0.426    &  0.250&  25.278  &      55.711    \\

ETP + GCN   &  0.800   &     0.764  &      0.664  &    0.525 &0.798   &     0.703   &     0.498   &   0.318  &  0.795   &     0.679   &     0.431 &     0.252&  33.486   &     80.479 \\
\midrule
\textbf{PG-GNN }  & \textbf{0.809}  &    \textbf{ 0.766} &      \textbf{0.665}    & \textbf{ 0.525} & \textbf{ 0.798}   &   \textbf{  0.707}  &      \textbf{0.502}  &   \textbf{ 0.318}   & \textbf{ 0.799 }   &  \textbf{  0.678 } &      \textbf{0.431}  &  \textbf{  0.253}&  \textbf{24.114}    &  \textbf{  37.688} \\
    \bottomrule
    \end{tabular}
\end{table*}

\begin{figure}[!h]
\centering
\subfloat[Dataset: CAMEO]{\includegraphics[width=3.5in]{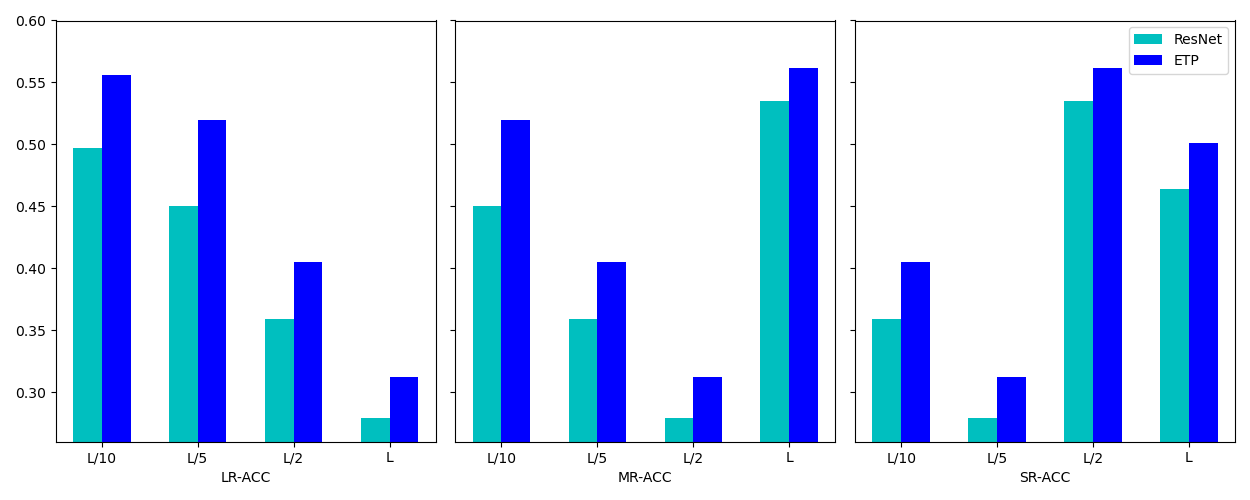}}
\hfil
\subfloat[Dataset: membrane]{\includegraphics[width=3.5in]{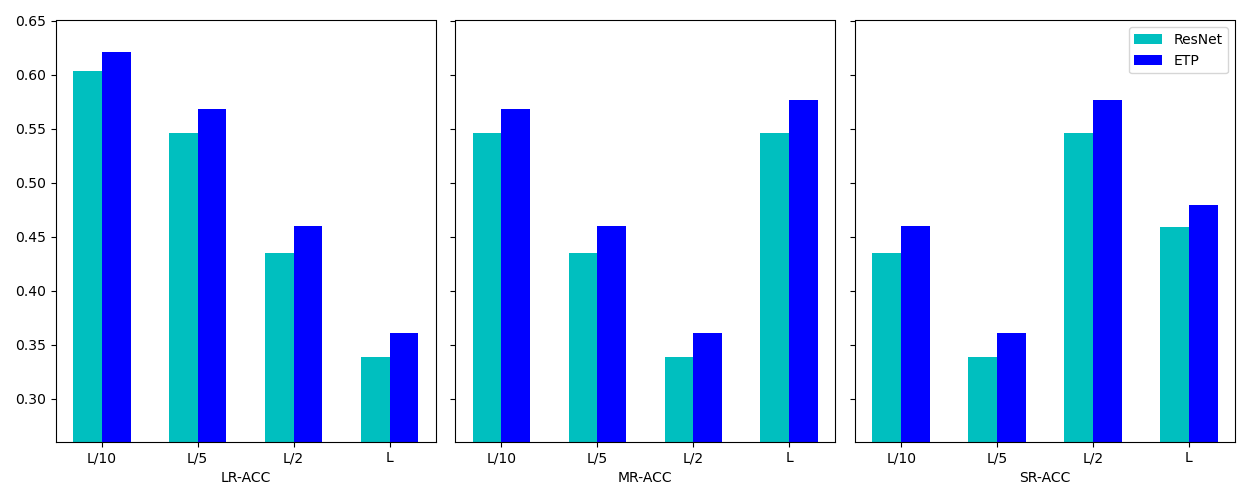}}
\hfil
\subfloat[Dataset: CASP13]{\includegraphics[width=3.5in]{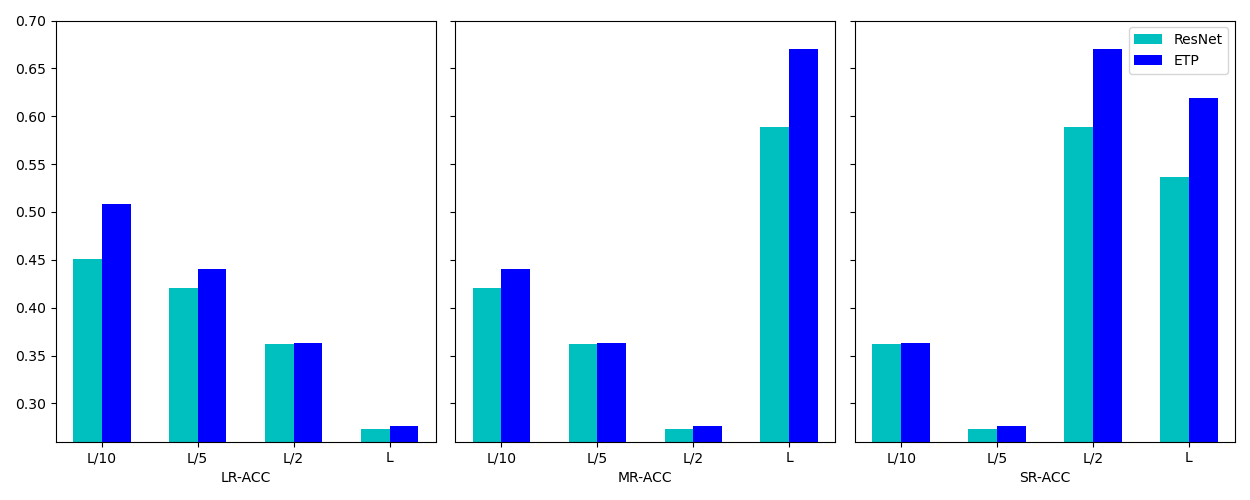}}
\hfil
\subfloat[Dataset: PDB25]{\includegraphics[width=3.5in]{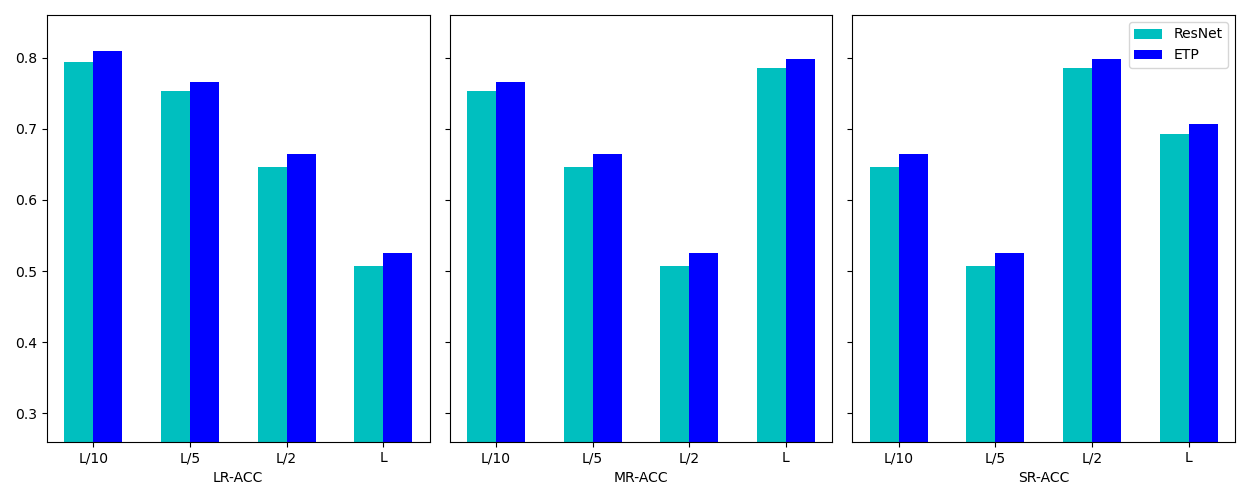}}
\hfil
\subfloat[Dataset: CASP11]{\includegraphics[width=3.5in]{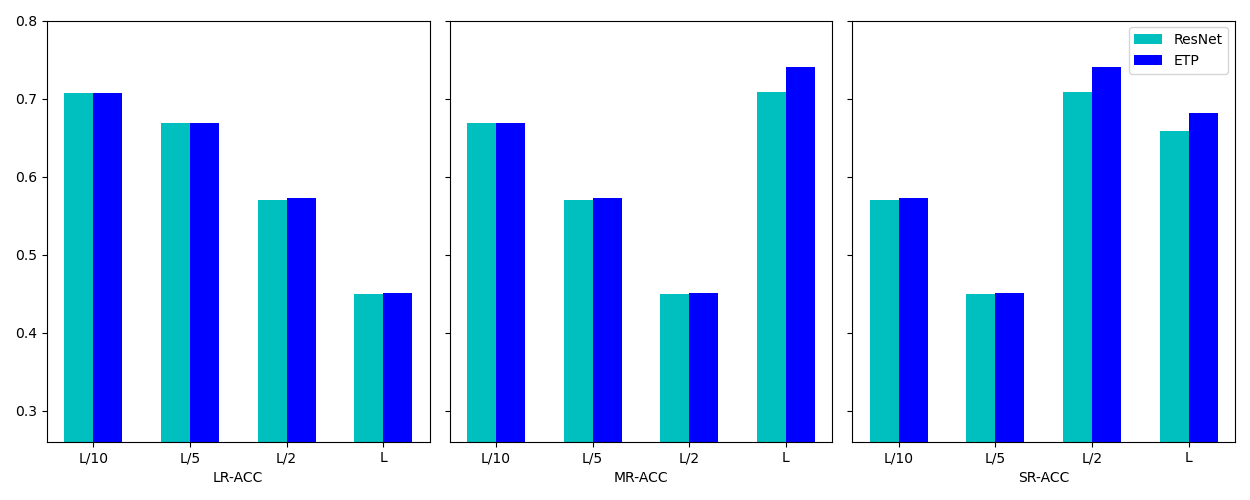}}
\caption{Performance comparison: proposed edge translation path (ETP) vs ResNet.
\\
}
\label{fig_e2e}
\end{figure}

\begin{figure}[!ht]
\centering
\subfloat[Dataset: PDB25]{\includegraphics[width=3.5in]{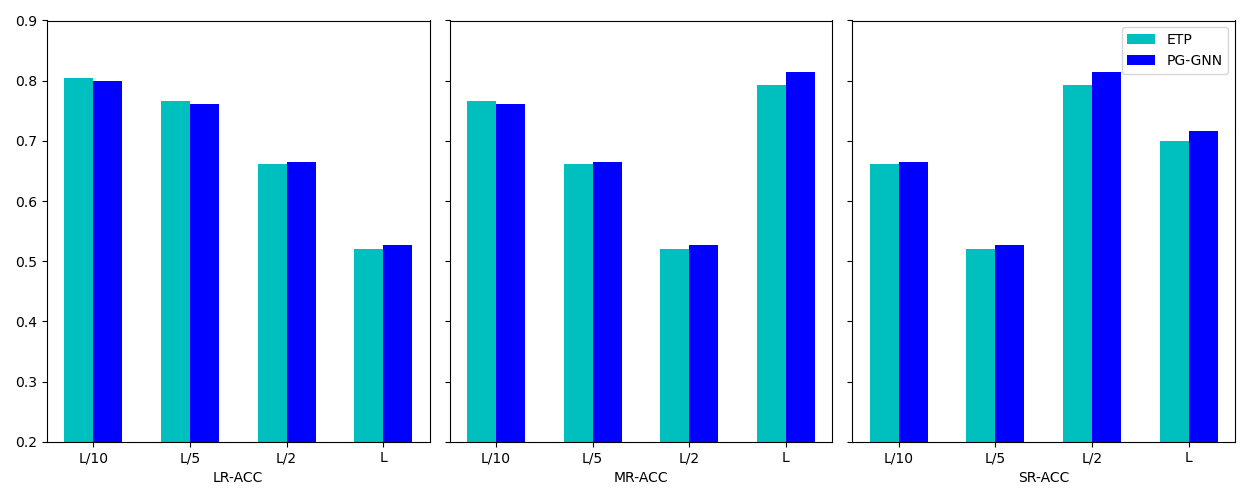}}
\hfil
\subfloat[Dataset: CAMEO]{\includegraphics[width=3.5in]{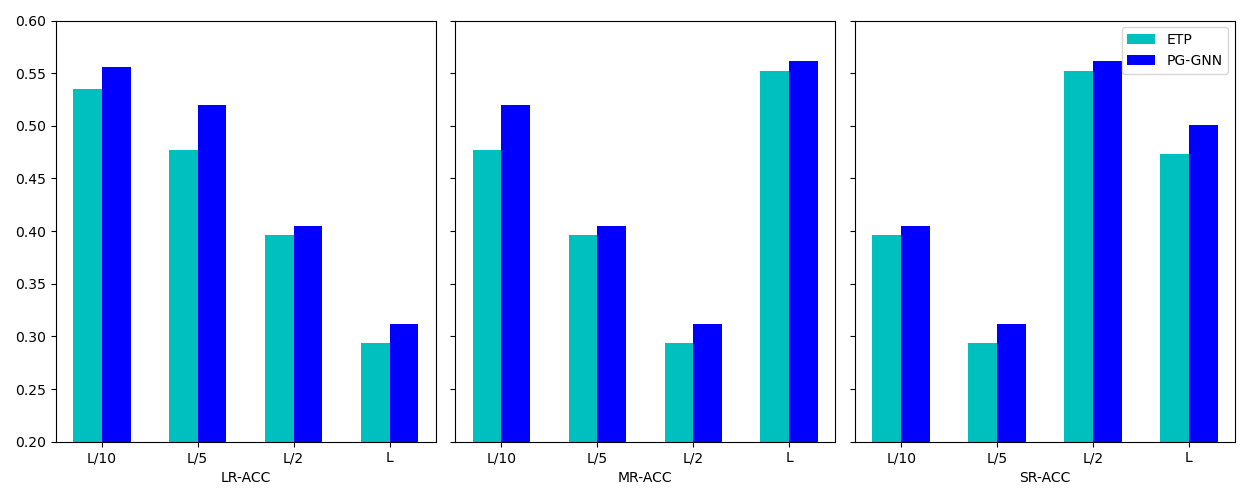}}
\hfil
\subfloat[Dataset: CASP11]{\includegraphics[width=3.5in]{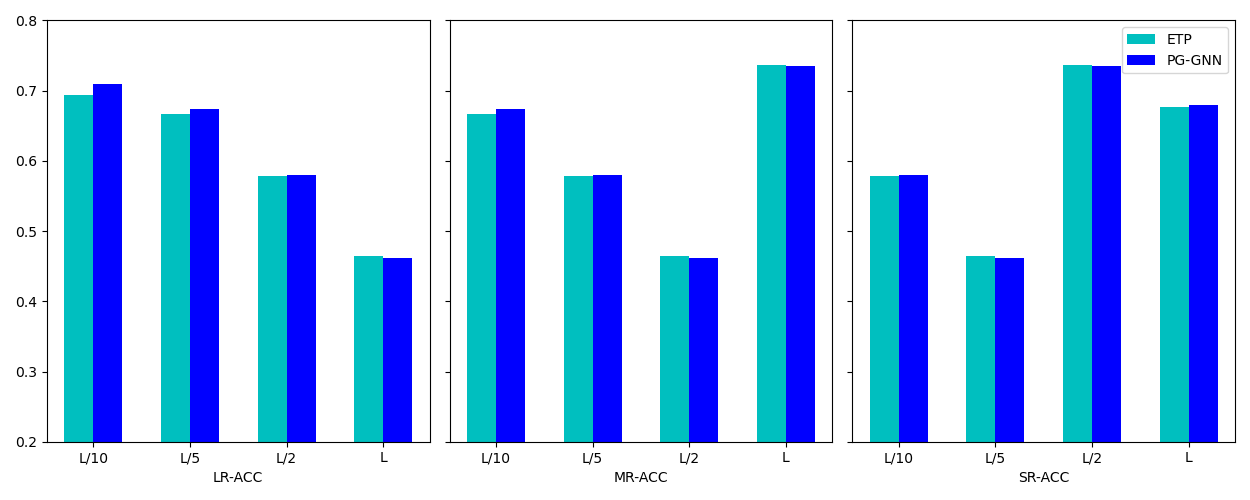}}
\hfil
\subfloat[Dataset: CASP13]{\includegraphics[width=3.5in]{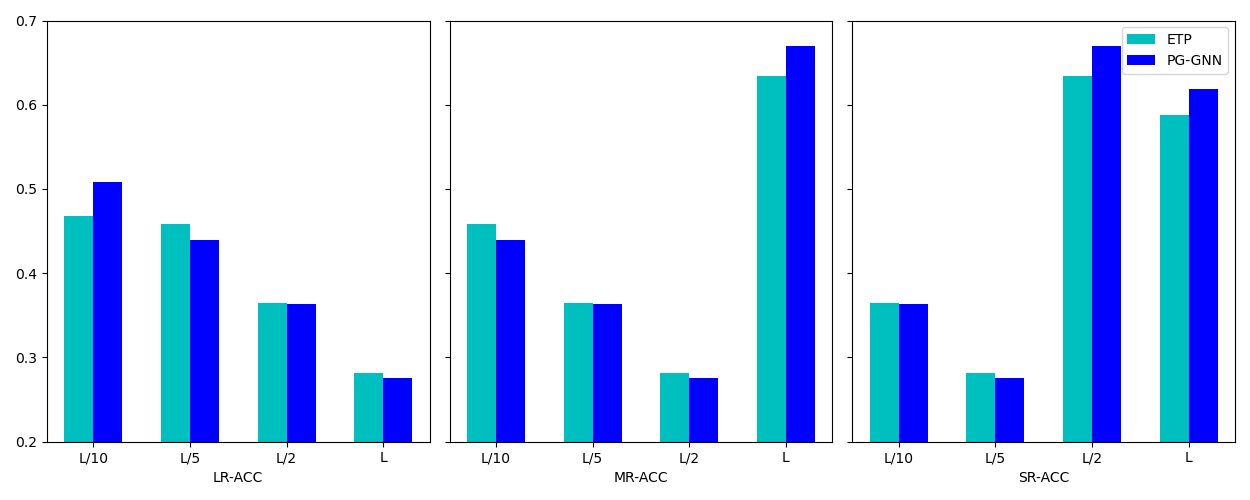}}
\hfil
\subfloat[Dataset: membrane]{\includegraphics[width=3.5in]{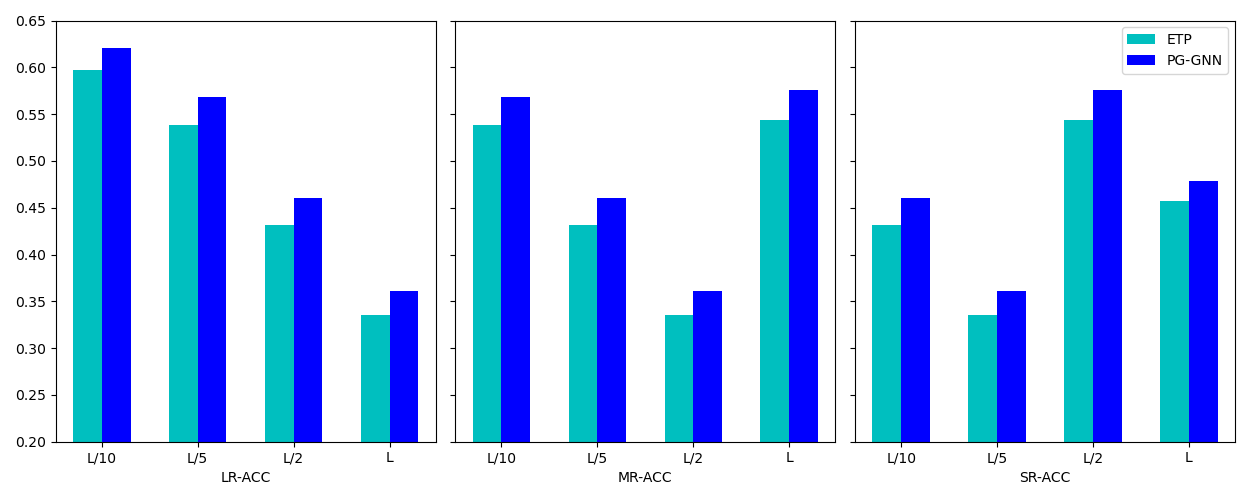}}
\hfil
\caption{Performance comparison: Our proposed framework PG-GNN vs Edge translation path only (ETP)}
\label{fig_proposed_framework}
\end{figure}

\begin{figure}[htbp]
\centerline{\includegraphics[width=\linewidth,height=\textheight,keepaspectratio]{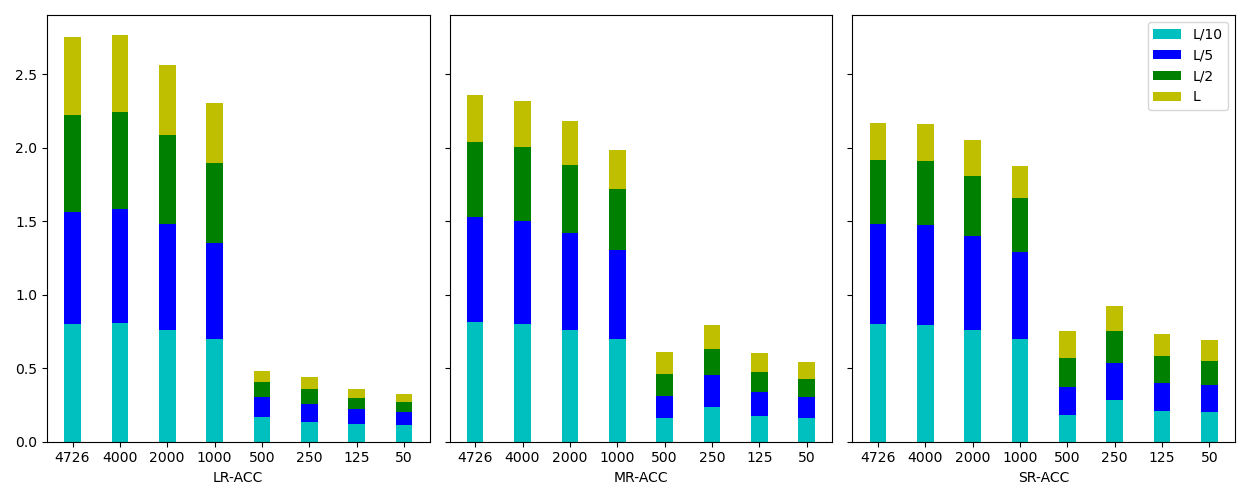}}
\caption{Sensitivity analysis of training dataset sizes }
\label{sensitivity analysis}
\end{figure}

\subsection{Benchmarks} \label{benchmarks}

As we are the first to propose a multi-task framework for protein graph representation learning,  our proposed method is compared with the two categories of methods: 
1) the methods used for protein edge (residue-residue contact) prediction only or protein node (torsion angles $(\phi, \psi)$) prediction only. We adapted the methods listed as follows. The methods were used in the proposed framework as baselines.  The methods were implemented to reproduce the results provided by the authors. The goal of re-implementing the methods was to investigate the performance with the same data. The existing web-servers provided by authors only make the inference based on the pre-trained model which is trained with different datasets. Hence, our re-implementation ensures baselines to train the models with the same data and features. Each method represents state-of-the-art performance in either contact map prediction task, torsion angle prediction, or graph node prediction task.

\begin{itemize}
    \item CCMpred: This method is proposed by \cite{ccmpred} using Markov Random Field pseudo-likelihood maximization for learning protein residue-residue contacts. The source code was directly used in the edge translation path of our framework as our baseline. 
    \item ResNet: This method was first adapted by \cite{Wang2017} to be used for protein contact map prediction. We re-implemented the neural network and used the same parameters stated in \cite{Wang2017} as our baseline for the edge translation path. 
    \item Dilated Residual Networks (DRN): This method was used by \cite{Senior2020ImprovedLearning}, \cite{badri2020}, for protein residue-residue contact prediction. We implemented the neural network and used the same parameters stated in Deepcon \cite{badri2020} as our baseline for edge translation path. 
    \item Graph Convolution Network (GCN): We adapted plain GCN with dynamic k-NN described in \cite{li2019deepgcns} as our baseline for node translation path. We used the same parameters as stated in \cite{li2019deepgcns} and K=5, 10, 20 for k-NN selection. 
    \item ETP: This is our proposed method for edge translation path. 
    \item Message passing neural network (MPNN): The method\cite{li2015gated} is described in the node translation path in our proposed framework. 
\end{itemize}

 2) We adapted the above mentioned methods into our proposed framework as baselines, namely CCMpred + GCN, CCMpred + MPNN, ResNet + GCN, ResNet + MPNN,  DRN + GCN,  DRN + MPNN, and ETP + GCN. 
 To further validate the effectiveness of the proposed node-edge joint convolution framework, we conduct comparison baseline models of node translation path and edge translation path trained separately as well. 

\subsection{Evaluation Metrics}

\subsubsection{Quantitative Evaluation}
A set of metrics are used to measure the similarity between the generated graph and the real graphs in terms of node and edge attributes. To measure the generated edge attribute performance, we follow the Critical Assessment of Structure Prediction (CASP) measurements\cite{CASP}. We use the accuracy (ACC) and error (ERR) of the top L/\textit{k} predicted contacts. L is the sequence length of nodes in each graph and \textit{k} = 10, 5, 2, 1. We evaluate the non-local contacts where the sequence distance belongs to the three groups: short range (SR) [6,11], medium range (MR) [12,23], and long range (LR) [24,$\infty$). To measure the generated node attributes performance, we use MAE (mean absolute error) between torsion attributes of generated and real target graphs.

\subsubsection{Qualitative Evalutaion}

Qualitative evaluation are performed to compare the reconstructed protein structure with the X-ray crystallograph protein structures by  the structure alignment of superimposed protein backbones\cite{PyMOL}. Full atom structures were constructed by PyRosetta full-atom relaxation with our generated distance and torsion angle restraints \cite{chaudhury2010pyrosetta}\cite{yang2020improved}. The lowest-energy full atom model was selected for quality evaluation and visualization. 

\section{Result}
\subsection{Investigating Performance of Multi-Attribute Properties}
The performances of the methods described in Section \ref{benchmarks} are shown in Fig. \ref{result_comparison}. The detailed results of the PDB25 dataset are listed in Table I. Fig. \ref{result_comparison} composed of five subplots, each illustrates the results of five different datasets. The datasets used in Fig. \ref{result_comparison} are membrane, CASP13, CAMEO, CASP11 and PDB25, respectively. The color of the points refers to the specific method listed on the top of Fig. \ref{result_comparison}. The values on the \textit{x}-axis show the average error rate of the models' performance for the edge translation path. Similarly, the values on the \textit{y}-axis show the average MAE of the models' performance for the node translation path. Because lower is better for both the average error rate and average MAE, the best performance model is the one that is closest to the (0,0) point. Our proposed PG-GNN (red cross) shows the best performance across all five datasets. PG-GNN proves the ability to handle both the geometric node attributes and geometric edge attributes together. The ResNet+MPNN shows a close performance with our proposed PG-GNN, which is because the ResNet+MPNN doesn't have the edge-to-edge convolution branch in the edge translation path. CCMPred was compared as a  co-evolutionary analysis method for the edge translation path. All the deep neural network models used in edge translation path performs at least 47.1\% better in predicting the edge representations than threading based method CCMPred. All of the proposed node translation models (MPNN) performs 25.2\% better than the GCN models on average. We did an ablation study in the next section to further investigate the performances of the proposed edge-to-edge convolution branch and the node translation path.

\subsection{Ablation Study}
\subsubsection{Evaluation of the proposed graph edge convolution block}
We compared the robustness of the proposed edge translation path with the edge translation path without the edge-to-edge convolution shown in Fig. \ref{fig_e2e}. The light blue bar shows the performance of the network without the edge-to-edge convolution branch. The dark blue bar shows the performance of our proposed framework. The results of all five datasets follow the same trend. We observe similar performance gains with the addition of the edge-to-edge convolution branch. The average ACCs with the proposed ETP for membrane, CASP13, CAMEO, CASP11 and PDB25 datasets are 45.5\%, 48.4\%, 44.8\%, 57.6\%, and 60.4\%,  while the average ACCs without the edge-to-edge convolution branch are 43.3\%, 44.3\%, 42.1\%, 56.4\%, and 59.3\%, respectively. The proposed framework with the edge-to-edge convolution branch outperformed the model without the edge-to-edge convolution branch in all long-range, medium-range, and short-range metrics. The highest average ACC improvement outperforms the baseline by 36.8\%. This proves that the proposed edge-to-edge convolution branch successfully helps the model's performance in learning edge-related representations.
\subsubsection{Evaluation of the node translation path}
Following the same approach used for the evaluation of the edge translation path, we compared the performance of our node translation path (MPNN) with GCN for translating node-level features in our proposed framework. We use \textit{k} = 5, 10, 20 for the training of GCN as the baseline. The results of average MAE for $(\phi, \psi)$ prediction compared with the ground truth are shown in Table II. The average MAE results show that our proposed node translation path successfully outperformed GCN models. Specifically, MPNN outperforms the comparison GCN methods by 20.6\%, 22.5\%, and 22.8\% on node attribute $\phi$ on average, and MPNN outperforms the comparison GCN methods by 48.4\%, 48.7\%, and 49.1\% on node attribute $\psi$.
Comparing with GCN, we show similar performance gains by combining our proposed node translation path with edge translation path baselines in Fig. \ref{result_comparison}. This proves that the superiority of the proposed node translation path in the interpretation of node-related representations.

\subsubsection{Evaluation of the joint convolution framework}
To evaluate the performance of the joint convolution framework, we compare our proposed joint convolution framework with a network that only contains our proposed edge translation path. The results are shown in Fig. \ref{fig_proposed_framework}. The light blue bar shows the performance of the network without the node translation path. The dark blue bar shows the performance of PG-GNN. 
The average ACC improvements of the joint convolution framework for dataset membrane, CASP13, CAMEO, CASP11, and PDB25 are 5.6\%, 3.6\%, 4.7\%, 0.9\%, and 1.2\%. 
The performance improvements are consistent across all five test datasets. The joint convolution framework performs better in almost all metrics than the edge translation path alone consistently. Thus, the proposed PG-GNN can not only jointly predict the node and edge attributes, but also performs better than the edge translation path alone.

\subsection{Sensitivity analysis on the effect of training dataset sizes}
To further evaluate the effect of dataset sizes, we perform a sensitivity analysis over eight different sizes of the training datasets. The full training dataset contains 4725 proteins from which we randomly sampled 4000, 2000, 1000, 500, 250, 125, and 75 proteins to train the models. Fig. 10 shows the prediction results of the PDB25 test dataset. The yellow, green, dark blue and light blue bar segment represents the average ACC results of the top L/10, L/5, L/2, and L predicted contacts. We use a stacked bar graph with each bar represents the specific size of the training dataset and colored bar segment represent different ranges' metrics. The performance of the network continues to increase with the increase in the size of the training dataset. We noticed a clear trend where the ACC increases marginally from 500 to 1000 proteins compared to other samples. 
The average ACCs of PG-GNN using 1000 proteins training set increased by 240\% comparing to 500 proteins training set. 
This trend from 500 to 1000 proteins fades when we go to larger datasets. The performances are consistent across all five test datasets.

\begin{figure*}[htbp]
\centering
\includegraphics[width=1.3in]{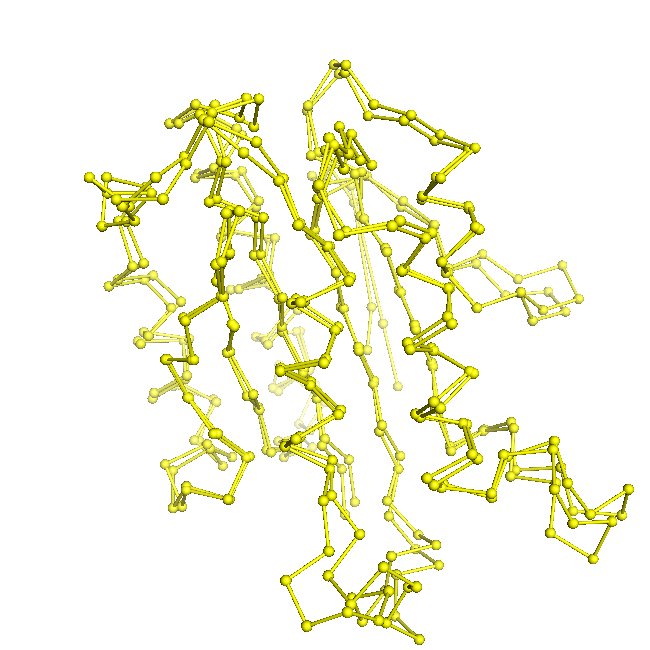}
\hfil
\includegraphics[width=1.3in]{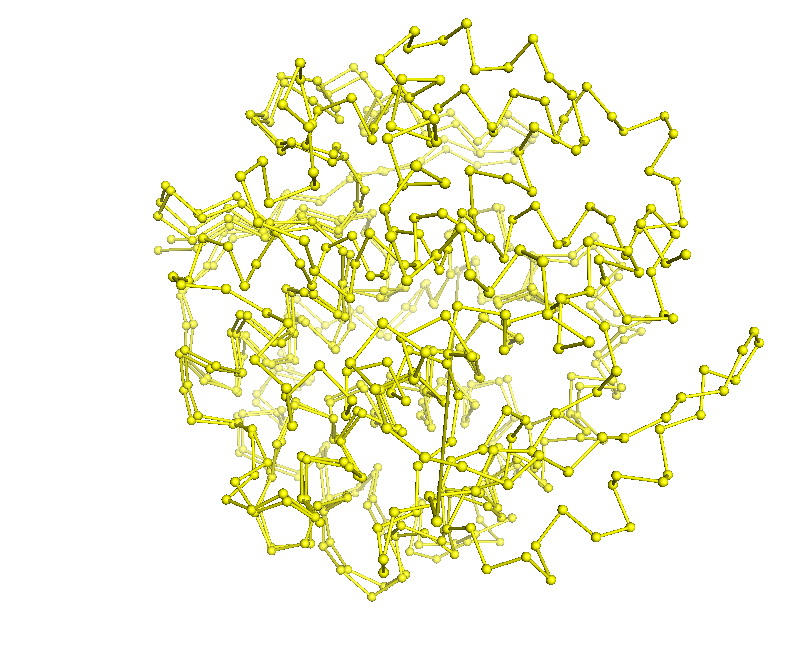}
\hfil
\includegraphics[width=1.3in]{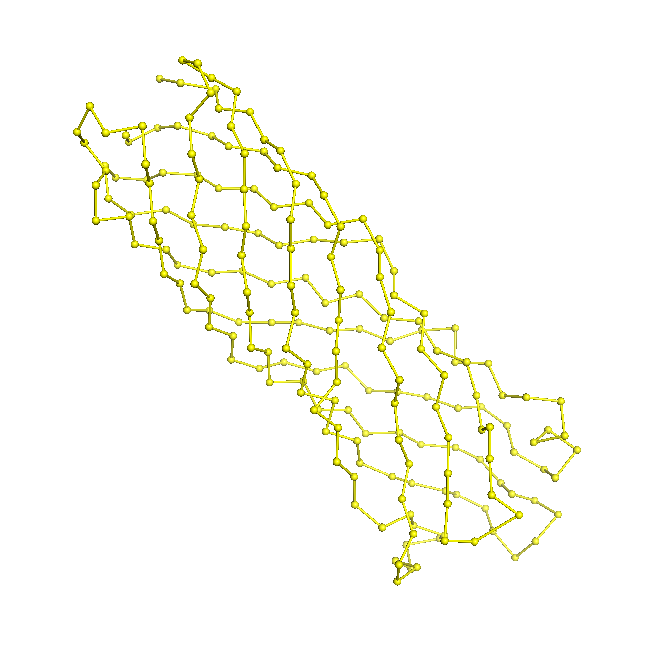}
\hfil
\includegraphics[width=1.3in]{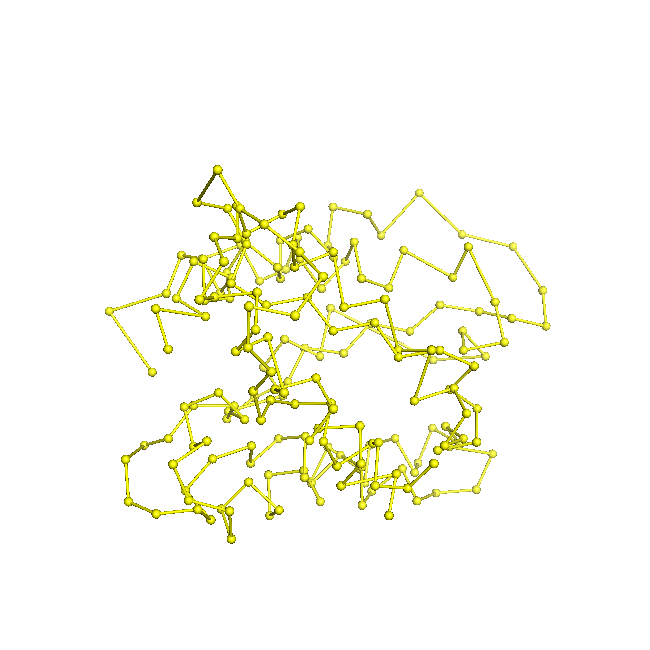}
\hfil
\includegraphics[width=1.3in]{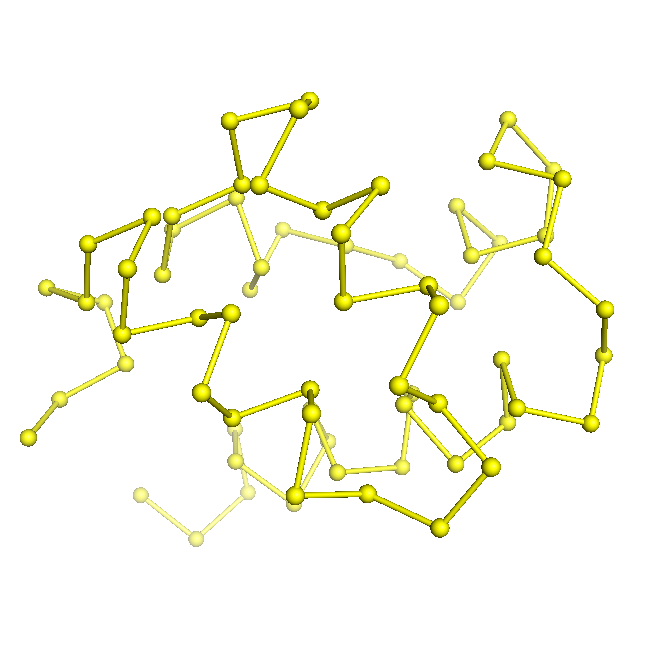}
\hfil
\subfloat[1q0p chain A]{\includegraphics[width=1.3in]{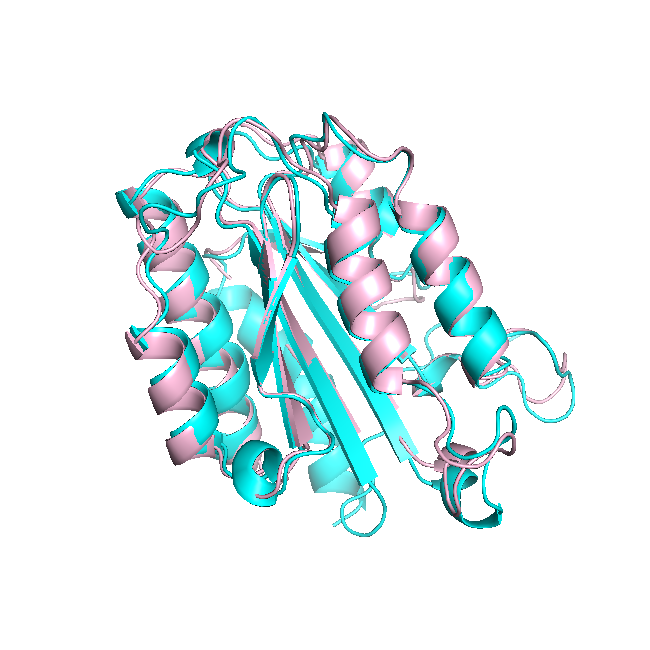}}
\hfil
\subfloat[2c4j chain A]{\includegraphics[width=1.3in]{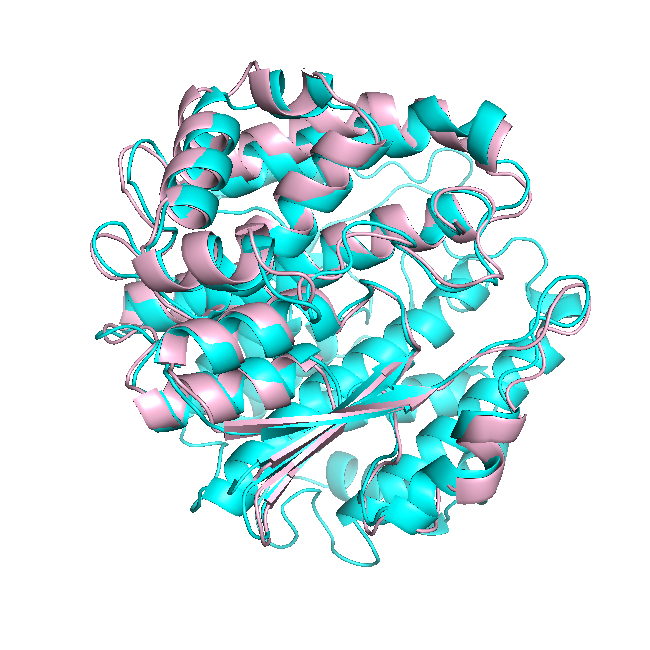}}
\hfil
\subfloat[2x27 chain X]{\includegraphics[width=1.3in]{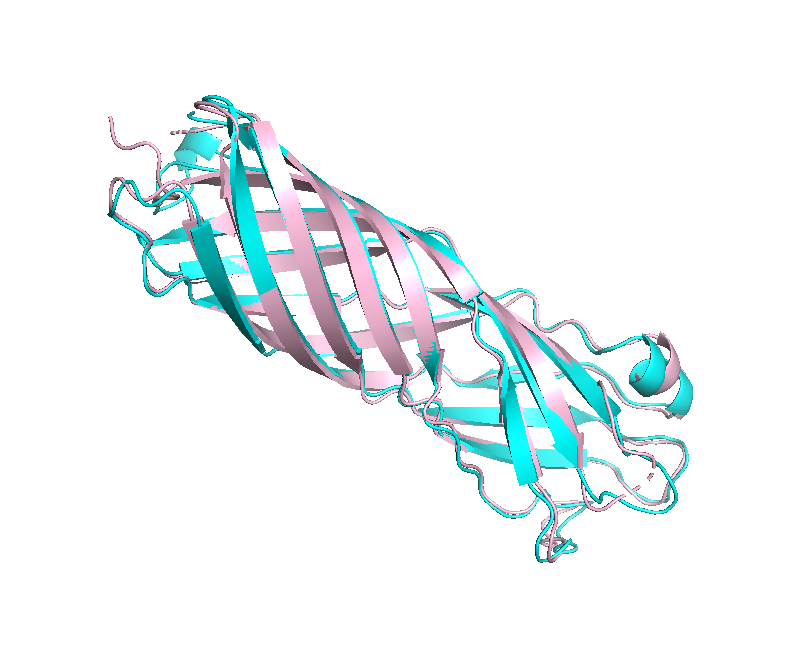}}
\hfil
\subfloat[3pb8 chain B]{\includegraphics[width=1.3in]{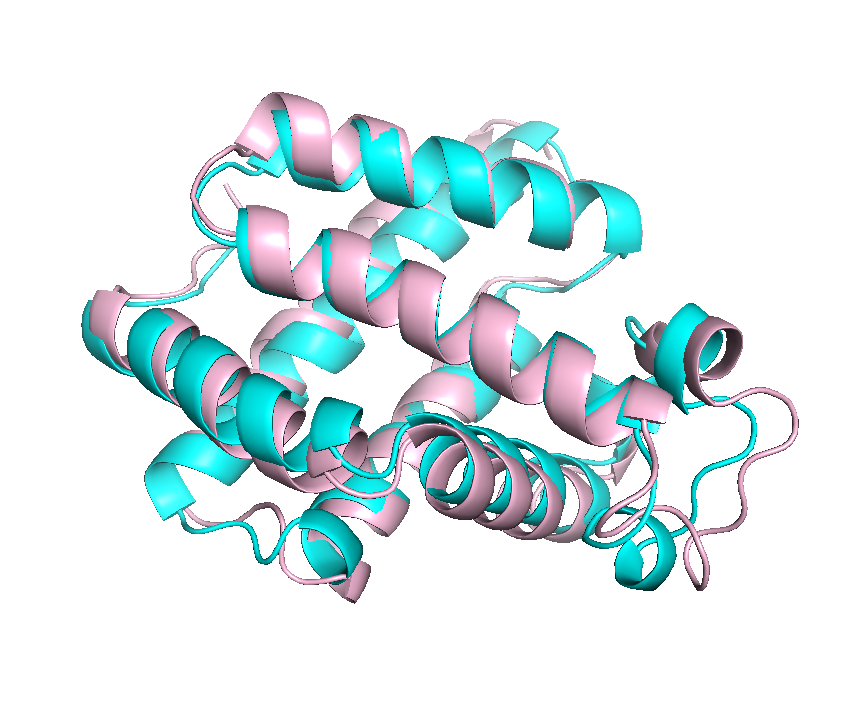}}
\hfil
\subfloat[4pzo chain A]{\includegraphics[width=1.3in]{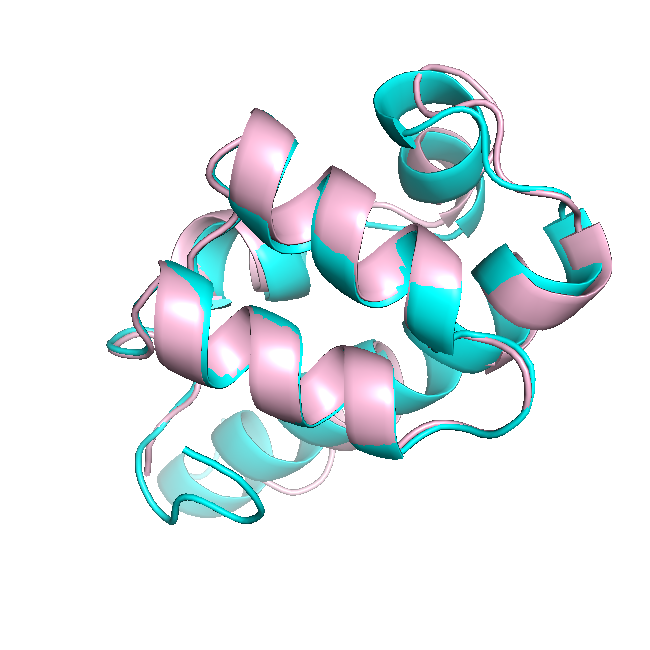}}
\hfil
\caption{The generated protein backbone trace(yellow), the superimposition of the predicted full atom model (light blue) and its native structure (pink).\label{superimpostition}}
\end{figure*}

\begin{table*}[htbp]
\label{node_comp}
  \centering
  \renewcommand{\arraystretch}{1.3}
  \caption{Performance comparison with different node translation path}
    \begin{tabular}{lccccccccccccccc}
    \toprule
     \multirow{2}[2]{*}{\centering Datasets}  & \multicolumn{2}{c}{GCN(k=5)} & \multicolumn{2}{c}{GCN(k=10)}& \multicolumn{2}{c}{GCN(k=20)}&\multicolumn{2}{c}{MPNN}\\ 
  \cmidrule(lr){2-3}   \cmidrule(lr){4-5}   \cmidrule(lr){6-7}  \cmidrule(lr){8-9}  & $\phi$  & $\psi$ & $\phi$  & $\psi$&  $\phi$  & $\psi$  &  $\phi$  & $\psi$ \\
        \midrule
CASP11  & 32.350     &   81.451   &       34.107    &    80.500  &  33.343  &      80.592  &   25.696 &  42.003  \\
CASP13 &   32.401   &     81.864       &    33.027   &     83.178 &32.876    &    84.556     &   28.729 &  46.944   \\
CAMEO  & 32.385    &    81.863         &    32.854    &    84.185   & 33.337    &    80.602    &   26.635 &  44.110  \\
membrane &  32.063    &    82.229    &     33.285    &    81.423 & 32.891    &    84.559     &   23.892 &  35.713    \\
PDB25  & 32.143  &      82.224        &  33.027   &     83.177 & 34.642     &   85.316      &    24.114 &  37.688   \\
  \bottomrule
    \end{tabular}%
\end{table*}%
\subsection{Case study: Quality evaluation of generated protein structure}
Fig. \ref{superimpostition} shows the case study of predicted structure representation of five domains using the visualization program PyMol \cite{PyMOL}. The predicted protein backbone traces were presented in the first row in yellow, which \ce{C_\alpha} atoms are balls and yellow lines between balls are backbone traces. The constructed full-atom model and the native structure are shown in the second row in cartoon representation. Our constructed models are light blue and their native structures are in pink. The constructed full atom models are consistent with the predicted backbone traces. The predicted structural models have very close topology as the native structures as shown in the figure. 

\section{Conclusion}
In this work, we revisit a long-existing challenge. Unlike treating the protein as a non-structural problem like an image, we propose to encode protein topological information into the node and edge representation in graphs. The experiments conducted in this paper proved the robustness of graph neural networks. Using our proposed model, we can achieve comparable results with less training time and smaller training datasets. As we establish an open problem for future research, it would be interesting to see if the performance keeps increasing with more features used. Because protein structures and inter-residue relations are ubiquitous in the real world, we left the further improvement of this longstanding challenge to future work.

\section*{Acknowledgment}
This research has been funded in part by the U.S. National Science Foundation grants IIS-1618669 (III) and ACI-1642133 (CICI).

\bibliographystyle{unsrt}
\bibliography{references}{}

\end{document}